\pdfoutput=1

\documentclass[11pt]{article}

\usepackage[final]{acl}

\usepackage{times}
\usepackage{latexsym}
\usepackage[T1]{fontenc}
\usepackage[utf8]{inputenc}
\usepackage{microtype}
\usepackage{inconsolata}
\usepackage{graphicx}
\usepackage{booktabs}
\usepackage{tabularx}
\usepackage{multirow}
\usepackage[table]{xcolor}
\definecolor{lightgray}{gray}{0.92}

\usepackage{array}
\usepackage{makecell}
\usepackage{amssymb} 
\usepackage{pifont}  
\usepackage{listings}

\usepackage[most]{tcolorbox}

\usepackage{xspace}

\usepackage{hyperref}
\usepackage{cleveref}

\usepackage[normalem]{ulem}

\newcommand{\name}{SATBench\xspace}

\definecolor{green}{RGB}{0,150,0}
\definecolor{red}{RGB}{220,50,50}

\newcommand{\cmark}{{\color{green}\ding{51}}}  
\newcommand{\xmark}{{\color{red}\ding{55}}}    

\newcommand{\datasetsize}{2100\xspace}

\newcommand{\humansize}{100\xspace}

\title{SATBench: Benchmarking LLMs' Logical Reasoning via Automated Puzzle Generation from SAT Formulas}

\author{
\textbf{Anjiang Wei}\textsuperscript{1}\thanks{Equal contribution.}\thanks{Correspondence to: anjiang@cs.stanford.edu}\hspace{2em} 
\textbf{Yuheng Wu}\textsuperscript{1}\footnotemark[1] \hspace{2em} 
\textbf{Yingjia Wan}\textsuperscript{2} \\
\textbf{Tarun Suresh}\textsuperscript{3} \hspace{2em} 
\textbf{Huanmi Tan}\textsuperscript{4} \hspace{2em} 
\textbf{Zhanke Zhou}\textsuperscript{1} \\
\textbf{Sanmi Koyejo}\textsuperscript{1} \hspace{2em} 
\textbf{Ke Wang}\textsuperscript{5} \hspace{2em} 
\textbf{Alex Aiken}\textsuperscript{1} \\
\\
\textsuperscript{1}Stanford University \hspace{1em} 
\textsuperscript{2}UCLA \hspace{1em} 
\textsuperscript{3}UIUC \hspace{1em} 
\textsuperscript{4}CMU \hspace{1em}
\textsuperscript{5}Nanjing University \\
}

\begin{document}
\maketitle

\begin{abstract}
We introduce SATBench, a benchmark for evaluating the logical reasoning capabilities of large language models (LLMs) through logical puzzles derived from Boolean satisfiability (SAT) problems.
Unlike prior work that focuses on inference rule-based reasoning, which often involves deducing conclusions from a set of premises, our approach leverages the search-based nature of SAT problems, where the objective is to find a solution that fulfills a specified set of logical constraints. Each instance in SATBench is generated from a SAT formula, then translated into a puzzle using LLMs. The generation process is fully automated and allows for adjustable difficulty by varying the number of clauses. All 2100 puzzles are validated through both LLM-based and solver-based consistency checks, with human validation on a subset. Experimental results show that even the strongest model, o4-mini, achieves only 65.0\% accuracy on hard UNSAT problems, close to the random baseline of 50\%. Our error analysis reveals systematic failures such as satisfiability bias, context inconsistency, and condition omission, highlighting limitations of current LLMs in search-based logical reasoning. Our code and data are publicly available at \url{https://github.com/Anjiang-Wei/SATBench}
\end{abstract}

\section{Introduction}
\label{sec:intro}

\begin{figure*}[!tb]
    \centering
    \includegraphics[width=\linewidth]{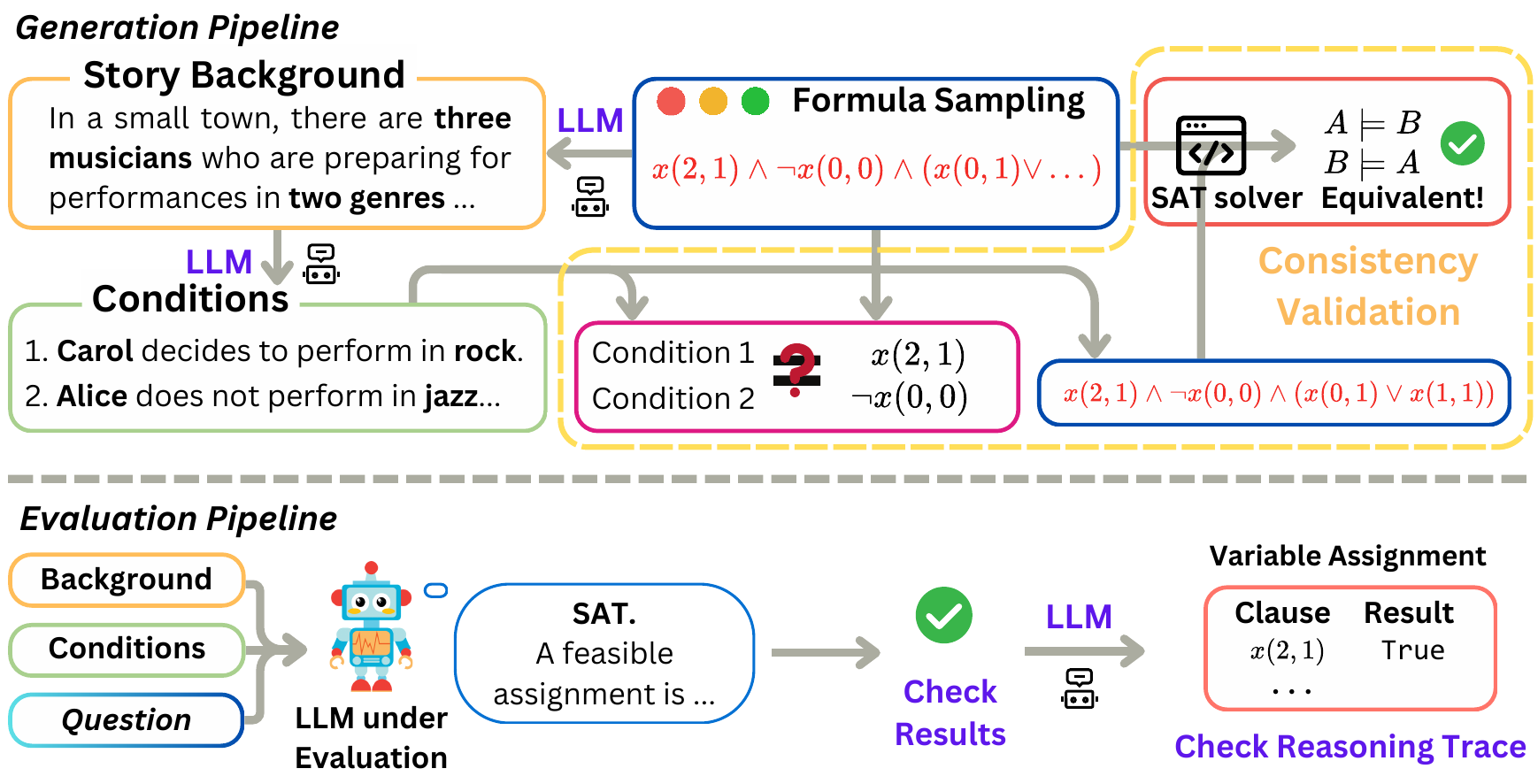}
    \caption{\textbf{Overview of the \name{} methodology.} The generation pipeline begins with sampling Conjunctive Normal Form (CNF) formulas, followed by LLM-driven creation of story backgrounds and conditions. To ensure the logical puzzle's quality, both LLM-assisted and solver-based consistency validations are employed. The evaluation pipeline then examines the puzzle's prediction outcomes and checks its reasoning process.}
    \label{fig:intro}
\end{figure*}

Logical reasoning is a fundamental component of human intelligence and continues to be a significant challenge in the field of artificial intelligence. The growing interest in the reasoning capabilities of large language models (LLMs) highlights the pressing need for robust benchmarks and evaluation methods~\cite{LogicGLUE}.

While many datasets have been proposed to evaluate logical reasoning capabilities of LLMs, earlier datasets do not exclusively evaluate logical reasoning in isolation, e.g., LogiQA~\cite{LogiQA}, and ReClor~\cite{ReClor}, which combine logical reasoning with commonsense reasoning.

Recently, new datasets have been introduced to assess logical reasoning in isolation, such as FOLIO~\cite{FOLIO} and P-FOLIO~\cite{P-FOLIO}. These datasets are manually curated by researchers and focus on logical problems based on \emph{inference rules}, which involve deriving conclusions from a set of premises.

\begin{table*}[htbp] 
\centering
\small
\resizebox{\textwidth}{!}{%
\begin{tabular}{@{}lccccccc@{}} 
\toprule
\textbf{Benchmark} & \textbf{\makecell{Search- \\ Based}} & \textbf{\makecell{Logic \\ Isolation}} & \textbf{\makecell{Automated \\ Generation}} & \textbf{\makecell{Difficulty \\ Control}} & \textbf{\makecell{Natural \\ Language}} & \textbf{\makecell{Template- \\ Free}} & \textbf{\makecell{Reasoning \\ Evaluation}}\\
\midrule
LogiQA~\cite{LogiQA}               &  \xmark  & \xmark & \xmark & \xmark & \cmark & \cmark & \xmark \\
BIG-bench~\cite{srivastava2022beyond}& \xmark & \xmark & \xmark & \xmark & \cmark & \cmark & \xmark \\
ReClor~\cite{ReClor}                &  \xmark & \xmark & \xmark & \xmark & \cmark & \cmark & \xmark \\
RuleTaker~\cite{RuleTaker}          &  \xmark & \cmark & \cmark & \cmark & \cmark & \xmark & \xmark \\
LogicNLI~\cite{LogicNLI}            & \xmark  & \cmark & \cmark & \xmark & \cmark & \cmark & \xmark \\
FOLIO~\cite{FOLIO}                  &  \xmark  & \cmark & \xmark & \xmark & \cmark & \cmark & \xmark \\
P-FOLIO~\cite{P-FOLIO}              & \xmark  & \cmark & \xmark & \xmark & \cmark & \cmark & \cmark \\
LogicPro~\cite{jiang2024logicpro}   &  \xmark & \xmark & \cmark & \xmark & \cmark & \cmark & \cmark \\
ZebraLogic~\cite{ZebraLogic}        &  \cmark & \cmark & \cmark & \cmark & \cmark & \xmark & \xmark \\
AutoLogi~\cite{AutoLogi}            &  \cmark & \cmark & \cmark & \cmark & \cmark & \cmark & \xmark \\
PARAT~\cite{pan2024can} & \cmark & \cmark & \cmark & \cmark & \xmark & \cmark & \cmark \\
LogicBench~\cite{LogicBench}                & \xmark     & \cmark & \cmark & \xmark & \cmark & \xmark & \xmark \\
LogicAsker~\cite{wan_2024_logicasker}             & \xmark     & \cmark & \cmark & \xmark & \cmark & \xmark & \xmark \\
Unigram-FOL~\cite{sileo_2024_scaling}                   & \xmark     & \cmark & \cmark & \xmark & \cmark & \xmark & \xmark \\
Multi-LogiEval~\cite{Multi-LogiEval}                   & \xmark     & \cmark & \cmark & \cmark & \cmark & \xmark & \xmark \\
\textbf{SATBench (ours)}                   & \cmark     & \cmark & \cmark & \cmark & \cmark & \cmark & \cmark \\
\bottomrule
\end{tabular}%
}
\caption{\textbf{Comparison of existing logical reasoning benchmarks.} An ideal evaluation framework should meet the following six criteria: (1) Logic Isolation: the benchmark exclusively evaluates logical reasoning in isolation; (2) Automated Generation: the benchmark construction is automated and scalable; (3) Difficulty Control: the difficulty levels of the benchmark questions are adjustable; (4) Natural Language: the questions are written in natural language rather than formal formulas; (5) Template-Free: the benchmark does not rely on expert-designed templates, enhancing diversity; (6) Reasoning Evaluation: the benchmark evaluates both the accuracy of model predictions and the correctness of their reasoning traces.}
\label{tab:compare}
\end{table*}

In this work, we introduce \name, a benchmark designed to create logical puzzles from Boolean satisfiability (SAT) problems~\cite{cook2023complexity,pan2024can} with LLMs. Unlike benchmarks based on inference rules, SAT problems are characterized as \emph{search-based} logical reasoning tasks, where the objective is to determine a truth assignment that fulfills a specified set of logical constraints~\cite{madusanka2024natural}. This approach to logical reasoning emphasizes a search process akin to backtracking used in SAT solvers. Unlike other search-based benchmarks such as ZebraLogic~\cite{ZebraLogic}, which presuppose the existence of a valid solution, SAT problems can result in either a satisfiable solution (SAT) or no solution (UNSAT).

As shown in \Cref{fig:intro}, starting from a SAT formula in Conjunctive Normal Form (CNF), such as \( (A \lor \neg B) \land (\neg C \lor \neg D) \), our framework uses LLMs to generate a story context and define a mapping between formula variables and entities in the story. Each clause is then translated into a natural language condition based on this mapping. By sampling CNF formulas with varying numbers of clauses, we can control puzzle difficulty. To ensure the quality of resulting logical puzzles, we reverse the generation process: LLMs translate the natural language conditions back into logical formulas, which are then compared to the originals using a combination of LLM-assisted and solver-based consistency checks. In the evaluation pipeline, we check the result and employ the LLM-as-a-judge strategy to assess the reasoning trace. To validate the overall process of story generation and reasoning trace evaluation, we manually checked \humansize examples, with passing rates above 90\%, which increases confidence in the quality of the dataset and evaluation protocol.

The evaluation on our generated \datasetsize logical puzzle dataset demonstrates that reasoning models exhibit strong performance on \name, with the o4-mini model achieving the highest accuracy. However, as the complexity of the problems increases, with a larger number of conditions in the logical puzzles, there is a noticeable decline in model performance. Specifically, the o4-mini model achieves an average accuracy of 65.0\% for the hard subset of UNSAT problems. This highlights the challenges posed by our benchmark, particularly for hard instances, where even the best-performing model only marginally surpasses the random baseline of 50\%, leaving significant room for improvement. Moreover, our analysis shows that while models often achieve higher accuracy on SAT than UNSAT problems, their reasoning traces are less reliable in the SAT setting, frequently predicting satisfiability without a valid assignment. To better understand these failures, we conduct an error analysis that identifies systematic patterns such as satisfiability bias, context inconsistency, and condition omission (\Cref{subsec:error}). These findings show that \name{} exposes limitations in current LLMs' ability to perform search-based logical reasoning. We further explore prompting and fine-tuning to improve performance on \name{} (\Cref{subsec:improving}). In summary, our work makes the following contributions:

\begin{itemize}
    \item \textbf{Task:} We present \name, a benchmark that uses large language models to generate logical puzzles from Boolean satisfiability (SAT) problems. The benchmark highlights the search-based nature of logical reasoning by focusing on finding truth assignments that satisfy given constraints.
    \item \textbf{Dataset:} Our generation process is fully automated and features adjustable difficulty levels by varying the number of clauses in SAT formulas. We ensure the quality of the \datasetsize generated logical puzzles through LLM-based and solver-based checks, with human validation showing passing rates above 90\%.
    \item \textbf{Analysis:} We show that accuracy declines sharply on the hard UNSAT problems and that reasoning traces for SAT are often unreliable. Our error analysis reveals systematic failures such as satisfiability bias, context inconsistency, and condition omission, highlighting limitations of current LLMs in search-based logical reasoning.
\end{itemize}

\section{Related Work}
\label{sec:related}

\paragraph{Logical Reasoning Benchmarks for LLMs}
Reasoning is a longstanding focus in NLP, with many benchmarks developed to assess model performance. Early efforts targeted natural language inference~\cite{bowman2015large} and commonsense reasoning~\cite{talmor2018commonsenseqa}, while recently there has been increasing attention to assessing logical reasoning, as seen in LogiQA~\cite{LogiQA}, ReClor~\cite{ReClor}, BoardgameQA~\cite{BoardgameQA}, and CLUTRR~\cite{CLUTRR}. These typically involve reasoning that relies on real-world knowledge. In contrast, datasets like FOLIO~\cite{FOLIO}, RuleTaker~\cite{RuleTaker}, and P-FOLIO~\cite{P-FOLIO} aim to isolate formal logical reasoning from commonsense knowledge. Logical puzzles have emerged as a compelling testbed in this area~\cite{giadikiaroglou2024puzzle}, with benchmarks including ZebraLogic~\cite{ZebraLogic}, AutoLogi~\cite{AutoLogi}, and LogicNLI~\cite{LogicNLI}. PARAT~\cite{pan2024can} examines LLMs directly on SAT formulas, whereas our benchmark frames SAT problems as natural language puzzles, a more realistic setting given LLMs' training on text and the availability of efficient SAT solvers for formula inputs. Unlike AutoLogi, which builds on existing corpora and risks data contamination, our dataset is generated entirely from scratch with solver and human validation to ensure correctness and diversity. A systematic comparison of these benchmarks is provided in \Cref{tab:compare}.

\paragraph{Logical Reasoning with Language Models}
Recent work investigates how large language models engage in logical reasoning via prompting techniques, supervised training on reasoning datasets, and translation into formal logic. A prominent line of research focuses on prompting methods that elicit step-by-step reasoning, including chain-of-thought prompting~\cite{wei2022chain}, tree-of-thought prompting~\cite{yao2023tree}, along with other methods~\cite{zelikman2022star,kojima2022large,li2022advance,tyagi2024step,chen2025finereason}. Another approach involves fine-tuning LLMs on datasets specifically designed for logical reasoning~\cite{abductionrules,FLD,LogicGLUE,dziri2023faith,ranaldi2024aligning}, which has demonstrated improved performance on formal reasoning benchmarks. Complementary to these methods, some work treats LLMs as semantic parsers that convert natural language reasoning tasks into formal logical representations, which are then executed or verified by external solvers or theorem provers~\cite{ye2023satlm,ryu2024divide}. In our evaluation, we use chain-of-thought prompting and prohibit models from invoking external tools; solvers are used only during dataset generation for validation.

\section{Method}
\label{sec:method}

\begin{figure*}[t]
    \centering
    \includegraphics[width=\linewidth]{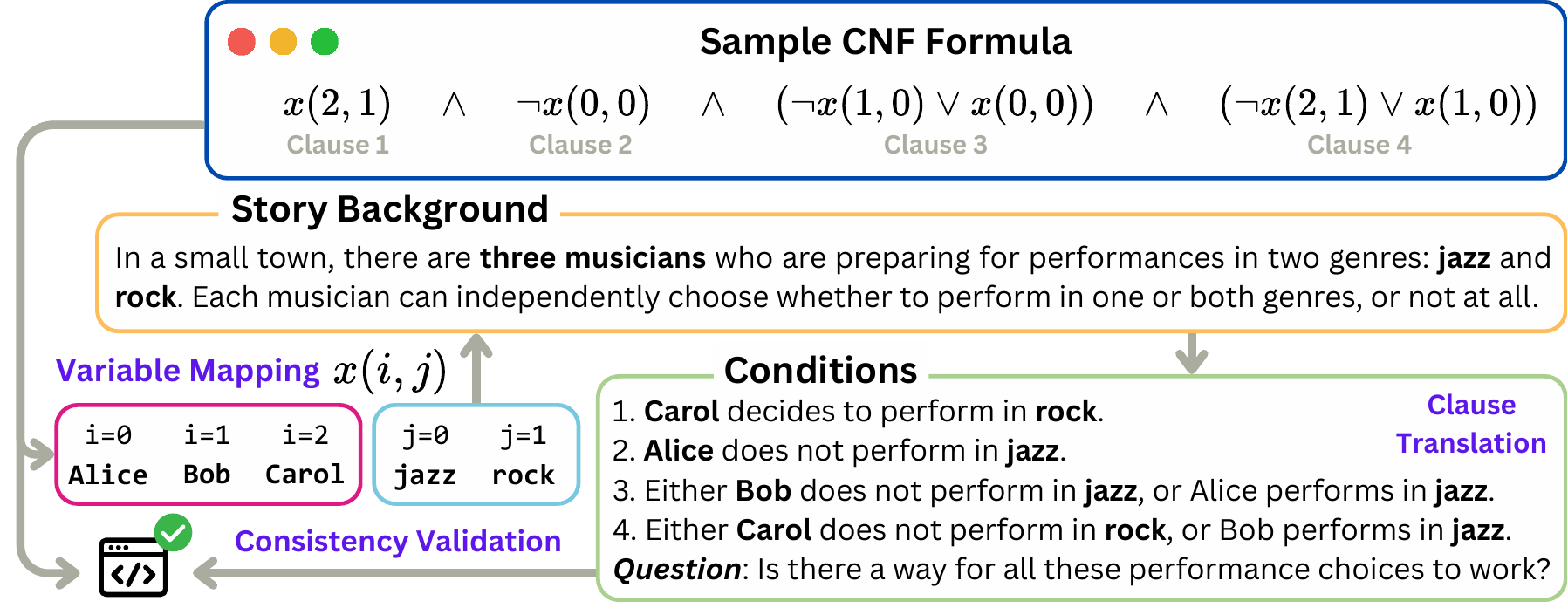}
    \caption{\textbf{Benchmark curation pipeline.} The process starts with sampling SAT formulas, followed by using an LLM to generate variable mappings and a story background. Clauses in the formula are then translated into narrative conditions. Consistency between the original formula and the generated puzzle is ensured through both LLM-based and solver-based validation.}
    \label{fig:method}
\end{figure*}

Our objective is to create logical puzzles derived from Boolean satisfiability (SAT) formulas, ensuring the quality of the dataset through both LLM-based and solver-based consistency checks. We further validate each LLM-involved process with human review. The generation method is divided into three stages: SAT formula sampling (\Cref{subsec:formula}), LLM-based story generation (\Cref{subsec:story}), and consistency validation (\Cref{subsec:consistency}). In the evaluation phase, we assess the correctness of the reasoning trace (\Cref{subsec:trace}).

\subsection{SAT formula Sampling}
\label{subsec:formula}

\paragraph{Conjunctive Normal Form (CNF)} Conjunctive Normal Form is a structured way of expressing logical formulas, where a formula is a conjunction (AND) of one or more disjunctions (OR) of literals. Each disjunction is referred to as a clause, and each clause consists of literals, which can be either a variable or its negation. For instance, the formula \((x(2, 1)) \land (\neg x(1, 0) \lor x(0, 0)) \land (\neg x(0, 0)) \land (\neg x(2, 1) \lor x(1, 0))\) is in CNF. Here, \(x\) represents a two-dimensional array with boolean elements, indicating true or false values. The SAT problem expressed in CNF form involves determining whether there exists an assignment of boolean values to the variables that satisfies the entire formula, making it true. If such an assignment exists, the formula is satisfiable. Conversely, if no such assignment can be found, the formula is unsatisfiable, and an UNSAT-Core can be identified, which is a subset of clauses that are inherently unsatisfiable. This approach constructs puzzles that challenge LLMs to determine if all conditions can be satisfied.

\paragraph{Automation and Difficulty Control}
The SAT problem can be solved using a SAT solver, which provides a soundness guarantee and allows for an automated and scalable solution. To systematically generate problems with varying levels of difficulty, we can sample formulas that differ in the number of boolean variables and clauses. Additionally, we can increase the dimensionality of the array to create more complex story contexts. By increasing the number of boolean variables, we can generate more clauses to be translated into story conditions. This approach effectively controls the difficulty level by expanding the search space and adding complexity to the constraints, making the search-based logical reasoning more challenging.

\subsection{Puzzle Story Generation}
\label{subsec:story}

\paragraph{Background and Variable Mapping}
To transform the sampled SAT formula into a narrative context, we utilize a language model, such as GPT-4o, to generate a story background and establish a mapping of variables. For example, as shown in \Cref{fig:method}, given the SAT formula, the language model creates a scenario involving three musicians: Alice, Bob, and Carol. These musicians are deciding on their performances in two musical genres, jazz and rock. Each musician can independently choose whether to perform in one or both genres, or not at all. The musicians and the genres correspond to the two dimensions of the array \(x\). This mapping is defined as:
\[
x(i, j) \rightarrow \text{``musician } i \text{ performs in genre } j\text{''}
\]
For example:
\begin{itemize}
  \item \( x(0,0) \): Alice performs in jazz
  \item \( x(1,0) \): Bob performs in jazz
  \item \( x(2,1) \): Carol performs in rock
\end{itemize}

\paragraph{Clause-to-Condition Mapping}
To transform each clause of the CNF formula into a narrative condition, we employ a large language model (e.g., GPT-4o). This transformation leverages the previously established story background and variable mapping. For example, the clause \(\neg x(0, 0)\) is translated to the condition ``Alice does not perform in jazz,'' while the clause \(\neg x(2, 1) \lor x(1, 0)\) is expressed as ``Either Carol does not perform in rock, or Bob performs in jazz.'' The final puzzle integrates the story background with these translated conditions and concludes with a question like ``Is there a way for all these performance choices to work?'' This question serves to assess the satisfiability of the conditions in the logical puzzle.

Our two-phase generation strategy, which begins with the creation of the story background and variable mapping, followed by the transformation of clauses into narrative conditions, improves the tractability and reliability of the process. This structured approach facilitates easier debugging and human validation. Additional examples of generated puzzles are provided in \Cref{app:puzzle-example}.

\subsection{Consistency Validation}
\label{subsec:consistency}

\paragraph{LLM-based Validation}
We utilize a large language model (GPT-4o) to ensure that each condition in the generated logical puzzle precisely matches the original SAT formula, given the specified variable mapping. This process checks that no extra conditions are introduced and none are missing. If the check fails, the puzzle is removed from our dataset. The validation prompt is provided in \ref{app:val-prompt}.

\paragraph{Solver-based Validation}
In addition to LLM-based checks, we adopt solver-based validation that enforces formula-level equivalence between the reconstructed formula and the original CNF. Using the variable mappings, an LLM first converts the narrative conditions back into a SAT formula. The original formula formula \(A\) and the reconstructed formula \(B\) are checked for equivalence using bidirectional entailment:
\[
A \equiv B \quad \text{iff} \quad A \models B \ \land\ B \models A.
\]
This condition is encoded into a SAT query and checked by the solver. Any reconstructed formula that fails equivalence is discarded, ensuring that the generated puzzles faithfully preserve the original logical structure.

\paragraph{Human Validation}
To ensure the quality of our dataset, we conduct human validation at two crucial stages involving LLMs, as detailed in \Cref{subsec:story}. The first stage involves the generation of the puzzle's background and variable mapping, where humans assess the logical coherence and confirm that the story background accurately reflects the independence of boolean variables. The second stage focuses on the translation of clauses into narrative conditions, where humans ensure that no additional constraints or misinterpretations are introduced.

\subsection{Reasoning Trace Evaluation}
\label{subsec:trace}
After generating the logical puzzles, we evaluate an LLM's performance using this dataset. Our evaluation emphasizes both the binary prediction result (SAT or UNSAT) and the validity of the model's reasoning trace. We adopt an LLM-as-a-judge methodology, where the model is instructed to produce a reasoning trace to justify its prediction. Below, we detail the approach for assessing the reasoning trace in SAT and UNSAT scenarios.

\paragraph{SAT Problems} When a problem is identified as SAT, it indicates that there is at least one assignment of True or False values to the variables that satisfies the CNF formula. Multiple solutions may exist. For example, consider the CNF formula \((x(0, 0) \lor \neg x(1, 0)) \land (x(1, 0) \lor x(2, 1))\). One possible satisfying assignment is \(x(0, 0) = \text{True}\), \(x(1, 0) = \text{False}\), and \(x(2, 1) = \text{True}\). After the model predicts a problem as SAT, it is required to generate a reasoning trace to support its prediction. We then instruct the judging LLM to translate this reasoning into a specific variable assignment using the given variable mapping. The judging LLM is further used to verify that each clause in the SAT formula evaluates to True, thereby confirming the satisfiability of the entire SAT formula.

\paragraph{UNSAT Problems} 
Unlike SAT problems, UNSAT problems have no variable assignment that satisfies all clauses. A SAT solver can identify an UNSAT-Core, which is a minimal subset of unsatisfiable clauses. When the model predicts UNSAT, it must provide a reasoning trace.

Consider the formula: \((x(2, 1)) \land (\neg x(1, 0) \lor x(0, 0)) \land (\neg x(0, 0)) \land (\neg x(2, 1) \lor x(1, 0))\). We can demonstrate its unsatisfiability through a step-by-step analysis:

\begin{enumerate}
    \item From the first clause, \(x(2, 1)\), we must set \(x(2, 1)\) to true.
    \item From the third clause, \(\neg x(0, 0)\), we must set \(x(0, 0)\) to false.
    \item Given that \(x(0, 0)\) is false, the second clause, \(\neg x(1, 0) \lor x(0, 0)\), can only be satisfied if \(\neg x(1, 0)\) is true, suggesting \(x(1, 0)\) is false.
    \item However, since \(x(2, 1)\) is true, the fourth clause, \(\neg x(2, 1) \lor x(1, 0)\), can only be satisfied if \(x(1, 0)\) is true.
\end{enumerate}

This leads to an irreconcilable contradiction: \(x(1, 0)\) is required to be both true and false simultaneously to satisfy all clauses, rendering the formula unsatisfiable. The example above illustrates a valid reasoning trace for an UNSAT problem in formula format. However, since the model being evaluated lacks access to the variable mapping during its reasoning trace generation, the judging LLM must first translate the reasoning trace back into the variable format. It then compares this translated reasoning with the provided UNSAT-Core to assess the accuracy of the reasoning trace.

\paragraph{Human Validation}
Given our use of an LLM-as-a-judge methodology for evaluating reasoning traces, we incorporate a human validation process to check the correctness of the LLM's judgments.
\section{Experimental Setup}
\label{sec:setup}

\paragraph{Dataset} The \name{} dataset consists of \datasetsize logical puzzle instances. Table~\ref{tab:dataset} provides statistics on the average number of boolean variables and clauses in the sampled SAT formulas, as well as the average number of words and sentences in the generated logical puzzles. The dataset generation process is fully automated, allowing for the creation of additional instances as required.

\begin{table}[!tb]
\centering
\small
\begin{tabular}{lcc}
\toprule
\textbf{Metric} & \textbf{Value} \\
\midrule
Number of Instances & \datasetsize \\
Average Number of Variables & 36.0 \\
Average Number of Clauses & 20.6 \\
Average Number of Words & 546.2 \\
Average Number of Sentences & 55.2 \\
\bottomrule
\end{tabular}
\caption{Dataset statistics for \name{}.}
\label{tab:dataset}
\end{table}

\begin{table*}[!tb]
\small
\setlength{\tabcolsep}{6pt}
\centering
\begin{tabular}{
    l|ccc|ccc|ccc|>{\columncolor{lightgray}}c
}
\toprule
\multirow{2}{*}{\textbf{Model}} 
& \multicolumn{3}{c|}{\textbf{SAT}} 
& \multicolumn{3}{c|}{\textbf{UNSAT}} 
& \multicolumn{3}{c|}{\textbf{Overall}} & \multirow{2}{*}{\cellcolor{white}\textbf{Avg.}}\\
& Easy & Medium & Hard & Easy & Medium & Hard & Easy & Medium & Hard \\
\midrule
\textit{Random Baseline}& \textit{50.0} & \textit{50.0}& \textit{50.0}& \textit{50.0}& \textit{50.0}& \textit{50.0}& \textit{50.0}& \textit{50.0}& \textit{50.0} & \textit{50.0}\\
LLaMA3.1-8B               & 57.9 & 60.0 & 48.9 & 30.4 & 14.8 & 17.5 & 44.1 & 37.4 & 33.2 & 38.2 \\
DeepSeek-Distill-7B       & 63.9 & 27.6 & 16.8 & 69.1 & 43.8 & 42.1 & 66.5 & 35.7 & 29.5 & 43.9 \\
Qwen3-1.7B                & 77.1 & 65.7 & 53.2 & 53.4 & 30.5 & 42.5 & 65.3 & 48.1 & 47.9 & 53.7 \\
gpt-4o-mini               & 82.1 & 82.4 & 90.7 & 42.3 & 12.9 & 13.2 & 62.2 & 47.6 & 52.0 & 53.9 \\
LLaMA4-Scout              & 84.3 & 76.7 & 66.4 & 52.0 & 24.3 & 37.5 & 68.1 & 50.5 & 52.0 & 56.9 \\
LLaMA3.1-70B              & 82.0 & 55.7 & 45.4 & 55.2 & 59.0 & 48.9 & 68.6 & 57.4 & 47.1 & 57.7 \\
gpt-4o                    & 85.5 & 83.3 & 78.6 & 54.3 & 27.1 & 18.9 & 69.9 & 55.2 & 48.8 & 58.0 \\
LLaMA3.3-70B              & 90.7 & 89.0 & 75.7 & 39.5 & 27.1 & 30.0 & 65.1 & 58.1 & 52.9 & 58.7 \\
DeepSeek-Distill-14B      & 82.9 & 51.4 & 41.1 & 85.7 & 59.0 & 51.8 & 84.3 & 55.2 & 46.4 & 62.0 \\
LLaMA4-Maverick           & 80.2 & 86.2 & 86.1 & 76.8 & 25.7 & 17.9 & 78.5 & 56.0 & 52.0 & 62.1 \\
Qwen3-4B                  & 84.1 & 78.1 & 78.6 & 80.7 & 31.9 & 22.1 & 82.4 & 55.0 & 50.4 & 62.6 \\
Qwen3-8B                  & 82.7 & 76.7 & 67.5 & 81.6 & 34.8 & 32.1 & 82.1 & 55.7 & 49.8 & 62.6 \\
DeepSeek-Distill-32B      & 84.5 & 53.8 & 42.1 & 90.0 & 68.1 & 58.6 & 87.2 & 61.0 & 50.4 & 66.2 \\
Qwen3-14B                 & 87.1 & 72.9 & 80.0 & 88.9 & 47.6 & 22.1 & 88.0 & 60.2 & 51.1 & 66.4 \\
Qwen3-235B-Int8           & 90.0 & 83.3 & 83.2 & 86.1 & 46.2 & 19.6 & 88.0 & 64.8 & 51.4 & 68.1 \\
Qwen-QwQ-32B              & 92.5 & 75.7 & 59.3 & 84.1 & 51.9 & 46.4 & 88.3 & 63.8 & 52.9 & 68.3 \\
Claude-3.7-Sonnet         & 88.4 & 77.6 & 83.6 & 93.8 & 63.3 & 42.1 & 91.1 & 70.5 & 62.9 & 74.8 \\
DeepSeek-V3               & 93.6 & 83.8 & 71.4 & 97.5 & 83.3 & 74.3 & 95.5 & 83.6 & 72.9 & 84.0 \\
DeepSeek-R1               & 94.8 & 87.1 & 73.6 & 98.2 & \textbf{89.5} & \textbf{83.6} & 96.5 & 88.3 & \textbf{78.6} & 87.8 \\
o4-mini                   & \textbf{97.0} & \textbf{96.7} & \textbf{91.1} & \textbf{98.2} & 88.1 & 65.0 & \textbf{97.6} & \textbf{92.4} & 78.0 & \textbf{89.3} \\

\midrule
Average                   & 84.1 & 73.2 & 66.7 & 72.9 & 46.4 & 39.3 & 78.5 & 59.8 & 53.0 & 63.8 \\
\bottomrule
\end{tabular}

\caption{Model accuracy on \name using zero-shot prompting for satisfiability prediction. Difficulty levels are categorized as follows: Easy (4-19 clauses), Medium (20-30 clauses), and Hard (31-50 clauses). All open-source models are instruction-tuned.}
\label{tab:main}
\end{table*}

\paragraph{Prompts}
We use 0-shot prompting to evaluate various LLMs on each logical puzzle in our dataset. Each puzzle's prompt includes a story background, a set of conditions that must be satisfiable simultaneously, and a query about their satisfiability. Models are required to generate a reasoning trace: if they determine the instance is satisfiable, they must provide a satisfying assignment for the variables; if they find it unsatisfiable, they must explain why the conditions cannot all be true at once. The final output must clearly state either \texttt{SAT} or \texttt{UNSAT}. Detailed prompts for the main evaluation and reasoning trace evaluation can be found in Appendix~\ref{app:prompt} and Appendix~\ref{app:trace-eval}, respectively.

\paragraph{Metrics}
In our evaluation, satisfiability is treated as a binary classification task, where random guessing results in a baseline accuracy of 50\%. The primary metric we use is the accuracy of the predicted satisfiability label. Besides, we also evaluate the correctness of the model's reasoning trace, but only if the satisfiability label is correct. We employ GPT-4o to determine whether the provided explanation logically supports the predicted outcome, as detailed in \Cref{subsec:trace}.

\paragraph{Models} We evaluate both proprietary and open-source language models. The proprietary models include GPT-4o~\cite{achiam2023gpt}, GPT-4o-mini, o4-mini, and Claude 3.7 Sonnet. The open-source models cover a range of recent ones from the Qwen~\cite{yang2025qwen3}, Llama~\cite{touvron2023llama}, and DeepSeek families~\cite{liu2024deepseek,guo2025deepseek}. For reasoning trace evaluation, we focus on the 5 top-performing models, and use GPT-4o as the judge.

\section{Results}
\label{sec:result}

\subsection{Main Results}

Table~\ref{tab:main} presents the accuracy on \name using zero-shot prompting for satisfiability prediction. Our findings are as follows.

\paragraph{Reasoning models excel in performance.} The o4-mini model stands out with a remarkable accuracy of 89.3\%. Close behind are the open-source models DeepSeek-R1 and DeepSeek-V3, with accuracies of 87.8\% and 84.0\%, respectively. Overall, reasoning models excel in our benchmark.

\paragraph{Model performance decreases with increasing problem difficulty.} We categorize difficulty levels as Easy (4-19 clauses), Medium (20-30 clauses), and Hard (31-50 clauses). Notably, even the top-performing model, o4-mini, sees its accuracy fall to 78.0\% on hard instances. Across all models, the average accuracy for hard problems is 53.0\%, which is nearly equivalent to the random baseline. More analysis of difficulty is provided in \Cref{subsec:difficulty}.

\paragraph{\name is a challenging benchmark.} For the hard instances, even the state-of-the-art model o4-mini only achieves 78.0\% accuracy, only a moderate improvement over the 50\% random baseline. For the UNSAT instances, its accuracy is only 65.0\%, leaving significant room for improvement.

\begin{figure}[!tb]
    \centering
    \includegraphics[width=\linewidth]{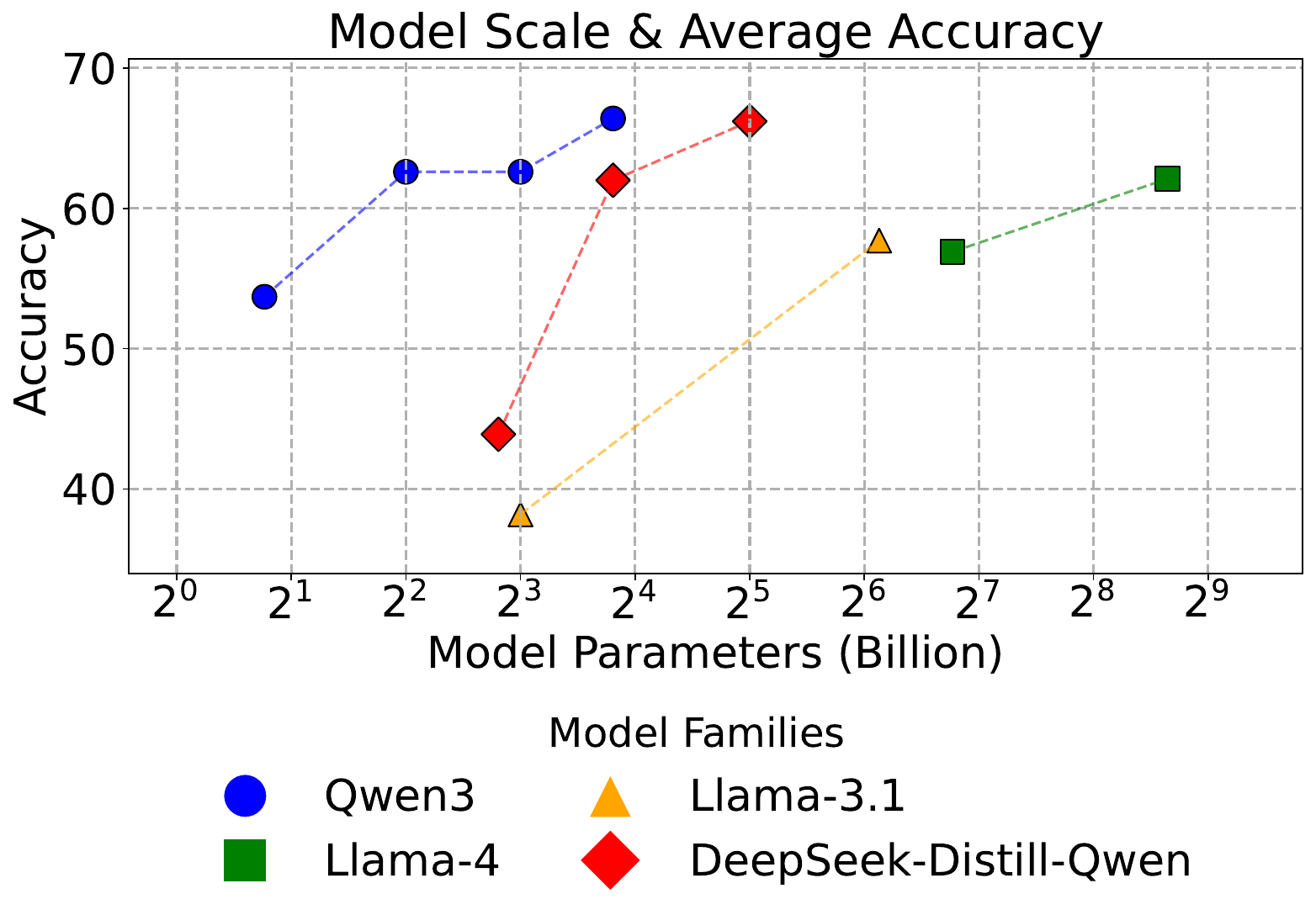}
    \caption{Scaling trend on \name.}
    \label{fig:scaling_trend}
\end{figure}

\paragraph{Scaling Trends.}
\Cref{fig:scaling_trend} shows that across model families such as Qwen3, Llama3.1, Mixtral, Llama4, and DeepSeek-Distill-Qwen, larger models generally achieve higher accuracy. Yet this trend does not hold uniformly across difficulty levels. On the hard instances, accuracy plateaus around 50–53\% even for the largest models in these families. Thus, the observed scaling gains are largely limited to easier problems. For the hardest cases, simply increasing model size yields little to no gain. These findings reinforce that SATBench remains a difficult and discriminative benchmark.

\begin{figure}[!tb]
  \centering
  \includegraphics[width=\linewidth]{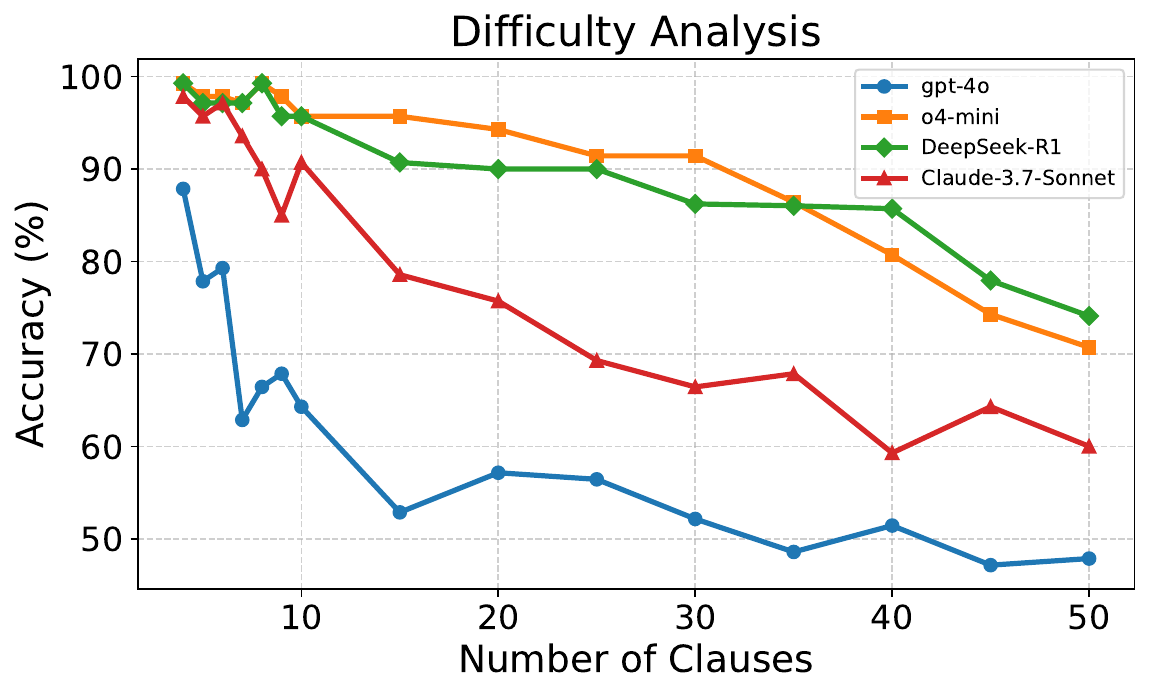}
  \caption{Impact of clause quantity on accuracy.}
  \label{fig:difficulty}
\end{figure}

\subsection{Analysis of Difficulty}
\label{subsec:difficulty}

\paragraph{SAT versus UNSAT}
The ``average'' row in \Cref{tab:main} highlights a notable disparity in model accuracy between SAT and UNSAT subsets. Models perform better on SAT problems, achieving an accuracy of 66.7\% on the hard instances, while only reaching 39.3\% on hard UNSAT problems. This suggests that SAT instances are generally easier for models to guess. Intuitively, the primary reason for this difference is that SAT is an existential property (there exists at least one satisfying assignment) while UNSAT is a universal property (all assignments fail to satisfy the formula).

\paragraph{Impact of Clause Quantity}
We examine the effect of the number of clauses on model accuracy. As shown in Figure~\ref{fig:difficulty}, there is a noticeable inverse relationship: model accuracy decreases as the clause count increases. For example, the GPT-4o model experiences a significant drop in performance, nearing random guess accuracy of 50\% as it approaches 30 clauses. This pattern suggests that a higher number of clauses adds complexity, demonstrating that our dataset generation methodology can effectively control difficulty levels.

\begin{table}[!tb]
\small
\centering
\begin{tabular}{c|cc|cc|c}
\toprule
  \multirow{2}{*}{\textbf{Model}} & \multicolumn{2}{c|}{\textbf{SAT}} & \multicolumn{2}{c|}{\textbf{UNSAT}} & \textbf{Overall} \\
  & Pred. & Trace & Pred. & Trace &  \textbf{Trace} \\ 
\midrule

QwQ  & 75.5 & 52.3 & 60.7 & 52.4 & 52.4\\
Claude-3.7 & 83.2 & 47.4 & 66.4 & 61.1 & 54.2  \\
DS-V3 & 82.9 & 65.7 & 85.0 & 71.1 & 68.4    \\
o4-mini    & 94.7 & 74.6 & 83.6 & 74.1 & 74.4  \\
DS-R1 &   85.2 & 73.8 & 90.3 & 82.1 & 78.0 \\
\bottomrule
\end{tabular}
\caption{Accuracy in prediction and reasoning trace evaluation.}
\label{tab:trace}
\end{table}

\subsection{Reasoning Traces and Error Analysis}
\label{subsec:error}

\paragraph{Trace Evaluation}
We evaluate the reasoning trace validity of various models with GPT-4o, and the results are shown in \Cref{tab:trace}. A notable observation is the disparity in trace accuracy between the SAT and UNSAT subsets. Models generally exhibit a more pronounced drop in trace accuracy on SAT problems compared to UNSAT ones. This suggests that higher prediction accuracy on SAT problems does not necessarily imply a valid variable assignment. Instead, models often show a bias toward predicting SAT outcomes without a valid assignment as evidence.  

\paragraph{Error Analysis}
To provide a deeper understanding of model failures, we conducted a qualitative error classification of incorrect predictions, defining four major error types:  

\begin{itemize}
  \item \textbf{Satisfiability Bias:} Models answer SAT but give an invalid assignment.
  \item \textbf{Context Inconsistency:} Models contradict their earlier reasoning, such as assigning conflicting values across steps.
  \item \textbf{Condition Omission:} Models ignore one or more conditions in the reasoning trace.
  \item \textbf{Spurious Priors:} Models introduce commonsense assumptions that are absent from the given constraints.
\end{itemize}

We used GPT-4o to automatically classify errors into these categories, and the distribution for two representative models is shown in \Cref{tab:error-analysis}. The observed patterns highlight core challenges in search-based logical reasoning, including failures in backtracking, difficulty in maintaining context, and reliance on prior knowledge rather than provided constraints. See examples in \Cref{app:error-examples}.

\begin{table}[!tb]
\centering
\small
\begin{tabular}{lcc}
\toprule
\textbf{Error Type} & \textbf{o4-mini (\%)} & \textbf{DS-R1 (\%)} \\
\midrule
Satisfiability Bias   & 68.7 & 40.5 \\
Context Inconsistency & 17.0 & 44.7 \\
Condition Omission    & 14.3 & 12.6 \\
Spurious Priors       & 0.0  & 2.1  \\
\bottomrule
\end{tabular}
\caption{Error type distribution in o4-mini and DeepSeek-R1.}
\label{tab:error-analysis}
\end{table}

\subsection{Human Validation}
\label{subsec:human}

We conducted human validation on a uniformly random sample of \humansize puzzles from our generated dataset to verify the correctness of LLM-involved steps and the reliability of our evaluation protocol. Each puzzle contains a CNF formula, its satisfiability label (SAT or UNSAT) from a symbolic solver, a narrative scenario with variable mappings, natural language conditions corresponding to each clause, a reasoning trace generated by an LLM, and an LLM judgment of whether the reasoning trace is logically valid. Three co-authors independently annotated the sample and resolved disagreements by majority vote.

Annotators performed three validation tasks:
\begin{enumerate}
    \item \textbf{Scenario and Mapping Consistency:} Ensuring that all entities in the scenario are covered in the variable mapping, and that every logical variable is correctly grounded. We observed no errors (100\% accuracy).
    \item \textbf{Clause Translation Faithfulness:} Verifying that each clause in the CNF formula is faithfully translated into its natural language condition without omissions, additions, or misinterpretations. We found minor translation errors in three cases, yielding a 97\% accuracy rate.
    \item \textbf{LLM Judgment Correctness:} Checking whether the LLM's judgment of the reasoning trace is logically correct and aligned with the ground-truth formula and satisfiability label. Here, accuracy was 93\%, with occasional errors due to incomplete assignment extraction or overly strict interpretations of valid traces.
\end{enumerate}

Overall, these results confirm the robustness of our dataset and evaluation pipeline, with errors being rare and not significantly affecting reliability.

A few failure cases were observed. In story generation, one error involved the clause $(\lnot x(2, 0) \lor x(2, 1))$ being translated as ``if Dr. Brown is not assigned project 0, then Dr. Brown is assigned project 1.'' This misuses the \textit{if-then} structure. The correct phrasing is ``if Dr. Brown is assigned project 0, then Dr. Brown is also assigned project 1.''

For the LLM-as-judge setting, the main error mode involved incomplete extraction of the assignment within the trace. In some cases, the model judged that the trace was invalid, even though the trace was logically sound. These minor errors, however, were rare and did not affect the overall robustness of our pipeline.
\section{Discussion}
\label{sec:discussion}

\subsection{SAT in Natural Language}

\name{} frames SAT problems as natural language puzzles rather than evaluating LLMs directly on SAT formulas. Testing only on symbolic inputs overlooks how reasoning arises in practice, since real-world tasks are almost always expressed in natural language. Because LLMs are trained mainly on text, narrative puzzles provide a more faithful and revealing evaluation of their reasoning ability. Additionally, our goal is not to replace SAT solvers, but to examine whether LLMs can reason about SAT structures when expressed in natural language, something classical solvers cannot address.

To test this distinction, we also evaluated models directly on CNF formulas. As shown in Table~\ref{tab:cnf-vs-puzzle}, accuracy was consistently higher on raw SAT inputs than on narrative puzzles, showing that natural language introduces additional complexity and makes the benchmark more challenging.

\begin{table}[!t]
\centering
\small
\begin{tabular}{lcc}
\toprule
\textbf{Model} & \textbf{Puzzle Acc. (\%)} & \textbf{Formula Acc. (\%)} \\
\midrule
GPT-4o-mini & 53.6 & 54.8 \\
GPT-4o      & 58.2 & 59.2 \\
DeepSeek-V3 & 84.0 & 87.3 \\
o4-mini     & 89.4 & 94.3 \\
\bottomrule
\end{tabular}
\caption{Comparison of model accuracy on narrative puzzles versus direct SAT Formula inputs. Natural language framing consistently increases task difficulty.}
\label{tab:cnf-vs-puzzle}
\end{table}

\subsection{Improving Performance on \name{}}
\label{subsec:improving}

\paragraph{Prompting}
Building on our error analysis in \Cref{subsec:error}, we designed error-aware prompts that explicitly remind models to avoid common pitfalls. Re-evaluating the previously misclassified cases under this setting led to substantial gains: for o4-mini, 60.4\% of failing cases were corrected, and for DeepSeek-R1, the rate is 73.2\%. These results show that making failure patterns explicit can significantly improve model performance.

\paragraph{Fine-tuning}
Using 1100 correct traces from o4-mini, we applied LoRA fine-tuning on Qwen2.5-14B-Instruct, raising accuracy from 51.9\% to 53.6\%. While modest relative to prompting, this indicates that supervised fine-tuning can help, with greater gains expected from larger datasets and reinforcement learning.

\section{Conclusion}

We present \name, a benchmark for assessing LLMs' logical reasoning via SAT-derived puzzles. Our dataset features search-based logical reasoning tasks, with controls difficulty and correctness checked by solvers and LLMs. \name{} contains 2100 logical puzzles, and we evaluate both satisfiability prediction and reasoning trace validity. Our findings show model performance drops with increased difficulty, with o4-mini scoring 65.0\% on the hard UNSAT cases, near the 50\% random baseline. We also conduct an error analysis that identifies systematic patterns such as satisfiability bias, context inconsistency, and condition omission. These findings show that \name{} exposes limitations in current LLMs' ability to perform search-based logical reasoning.

\section*{Limitations}
This paper utilizes LLMs, such as GPT-4o, for the generation of logical puzzles and consistency validation. While LLMs can enhance the scalability and diversity of our dataset, they could introduce potential inaccuracies that we cannot fully eliminate. To address this issue, we incorporate human validation to ensure a high-quality dataset. However, despite these efforts, the possibility of errors remains.

Another limitation of this work is its exclusive focus on the Boolean satisfiability problem for logical reasoning, which means that other forms of logical reasoning are not addressed by \name.


\section*{Acknowledgments}
We thank Yuan Liu, Xiaohan Wang, and Gabriel Poesia for their discussions. This work was partially
supported by a Google Research Award.

\bibliography{custom}

\setcounter{figure}{0}
\renewcommand{\thefigure}{A\arabic{figure}}
\setcounter{table}{0}
\renewcommand{\thetable}{A\arabic{table}}

\clearpage
\appendix
\section*{Appendix}

\label{sec:appendix}
\section{Example of Generated Puzzles}
\label{app:puzzle-example}
Please see \Cref{fig:puzzle-example}.

\begin{figure*}[t] 
\centering
\begin{tcolorbox}[title=Puzzle Example,
  colback=gray!5, colframe=gray!50,
  boxrule=0.8pt, arc=4pt, outer arc=4pt, breakable,
  width=\textwidth]  

\begin{lstlisting}[basicstyle=\ttfamily\small, breaklines=true, mathescape=true]
<SAT formula>
($\neg$x(0, 0) $\lor$ x(1, 0)) $\land$ (x(0, 1)) $\land$ ($\neg$x(1, 0)) $\land$ ($\neg$x(0, 1) $\lor$ x(1, 1)) 
$\land$ ($\neg$x(1, 1) $\lor$ x(0, 0))
</SAT formula>

<satisfiable>
false
</satisfiable>

<UNSAT reason>
Frozen conflict chain: sequential forced assignments leading to contradiction: 
(x(0, 1)), ($\neg$x(0, 1) $\lor$ x(1, 1)), ($\neg$x(1, 1) $\lor$ x(0, 0)), 
($\neg$x(0, 0) $\lor$ x(1, 0)), ($\neg$x(1, 0))
</UNSAT reason>

<scenario>
Two wildlife researchers, Hannah and Liam, are documenting animal behavior at a sanctuary. They are independently recording whether they observe two specific behaviors: feeding (0) and social interaction (1). Each researcher decides on their own which behavior they have observed, and they may report multiple behaviors or none at all.
</scenario>

<variable_mapping>
Let x(i, j) mean researcher i observes behavior j. 
Here, researcher 0 is Hannah, and researcher 1 is Liam.
</variable_mapping>

<conditions>
1. Either Hannah does not observe feeding, or Liam observes feeding.
2. Hannah observes social interaction.
3. Liam does not observe feeding.
4. Either Hannah does not observe social interaction, or Liam observes social interaction.
5. Either Liam does not observe social interaction, or Hannah observes feeding.
</conditions>

<question>
Is there a way to assign observations that make this work?
</question>

\end{lstlisting}

\end{tcolorbox}
\caption{Puzzle Example.}
\label{fig:puzzle-example}
\end{figure*}

\section{Templates}

\subsection{LLM Validation Prompt Template}
\label{app:val-prompt}
Please see \Cref{fig:val-prompt}.

\begin{figure*}[t] 
\centering
\begin{tcolorbox}[title=LLM Validation Prompt Template,
  colback=gray!5, colframe=gray!50,
  boxrule=0.8pt, arc=4pt, outer arc=4pt, breakable,
  width=\textwidth]  

\begin{lstlisting}[basicstyle=\ttfamily\small, breaklines=true, mathescape=true]
You are a logic checker.

You are given a SAT formula, a variable explanation, and a natural language puzzle based on the formula. Your job is to check whether the natural language conditions are logically equivalent to the original SAT formula.

Specifically, for each clause in the SAT formula:
- Verify there is a corresponding natural language condition with equivalent logical meaning.
- Ensure the variable usage matches the explanation format.
- Make sure there are no missing clauses, no added constraints, and no changes in logic.

Pay special attention to logical implications and how they are expressed in natural language. For example:

The clause ($\neg$x(2) $\lor$ x(1)) is logically equivalent to: ``If x(2) is true, then x(1) is also true." A common mistake is to write this as: ``If $\neg$x(2) then x(1)", which is incorrect. That corresponds to the clause (x(2) $\lor$ x(1)), and changes the meaning.

Here is the information:

<scenario>
{scenario}

<variable explanation>
{variable_mapping}

<conditions>
{conditions}

<question>
{question}

<SAT formula>
{formula}

Think step by step about whether the SAT formula and the natural language conditions match logically, clause by clause. Consider the number of clauses, the variable usage, and the logical operators involved. 

Your job is only to evaluate whether each condition correctly represents its corresponding clause in the SAT formula. You should not judge whether the overall formula or the scenario is satisfiable, solvable, or logically consistent.

Do not attempt to rewrite, fix, or invent any missing conditions. If any clause is missing, mistranslated, or not clearly represented, you must mark the result as [INCONSISTENT].

Finally, in the last line, output either [CONSISTENT] or [INCONSISTENT]. 
Do not include anything after this label.
\end{lstlisting}

\end{tcolorbox}
\caption{LLM Validation Prompt Template.}
\label{fig:val-prompt}
\end{figure*}

\subsection{SAT/UNSAT Evaluation Prompt Template}
\label{app:prompt}

Please see \Cref{fig:prompt}.

\begin{figure*}[t] 
\centering
\begin{tcolorbox}[title=SAT/UNSAT Evaluation Prompt Template,
  colback=gray!5, colframe=gray!50,
  boxrule=0.8pt, arc=4pt, outer arc=4pt, breakable,
  width=\textwidth]  

\begin{lstlisting}[basicstyle=\ttfamily\small, breaklines=true, mathescape=true]
You are a logical reasoning assistant. You are given a logic puzzle.

<scenario>
{scenario}

<conditions>
{conditions}

<question>
{question}

Guidelines:
- All constraints come **only** from the <conditions> section.
- The <scenario> provides background and intuition, but **does not impose any additional rules or constraints**.
- All variables represent **independent decisions**; there is no mutual exclusivity or implicit linkage unless stated explicitly in <conditions>.
- Variables not mentioned in <conditions> are considered unknown and irrelevant to satisfiability.

Your task:
- If the puzzle is satisfiable, propose one valid assignment that satisfies all the conditions.
- If the puzzle is unsatisfiable, explain why some of the conditions cannot all be true at once.

Think step by step. At the end of your answer, output exactly one of the following labels on a new line:
[SAT] - if a valid assignment exists  
[UNSAT] - if the constraints cannot be satisfied  

Do not add any text or formatting after the final label.
\end{lstlisting}

\end{tcolorbox}
\caption{SAT/UNSAT Evaluation Prompt Template.}
\label{fig:prompt}
\end{figure*}

\subsection{Trace Evaluation Prompt Template}
\label{app:trace-eval}

Please see \Cref{fig:trace-eval} and \Cref{fig:trace-eval2}.

\begin{figure*}[t] 
\centering
\begin{tcolorbox}[title=Trace Evaluation Prompt Template for SAT Prediction,
  colback=gray!5, colframe=gray!50,
  boxrule=0.8pt, arc=4pt, outer arc=4pt, 
  width=\textwidth]  

\begin{lstlisting}[basicstyle=\ttfamily\small, breaklines=true, mathescape=true]
You are given a logical puzzle and a reasoning trace from a language model.

The puzzle is also expressed as a SAT formula. Each clause is a disjunction (OR) of literals formatted like x(i), x(i,j), or x(i,j,k). These variables follow the meaning:

- x(i) means object or person i has some unnamed property.
- x(i,j) means object i has property or role j.
- x(i,j,k) means object i has property j in context or slot k (e.g., time, situation, location).

A positive literal like x(0,1) means that the property is present.  
A negative literal like $\neg$x(0,1) means it is absent.

Below is the full logical puzzle and its corresponding formula:

<scenario>
{scenario}

<conditions>
{conditions}

<final question>
{question}

<variable explanation>
{variable_mapping}

<SAT formula>
{formula}

<trace from model>
{model_trace}

Your task is to extract the truth assignment implied by the model's reasoning trace, and evaluate whether each clause in the SAT formula is satisfied.

Go through the trace and determine whether each variable appearing in the SAT formula is marked as True or False.

Then, for each clause, evaluate the truth value of each literal using this assignment.

For example, if a clause in the SAT formula is (x(0) $\lor$ $\neg$ x(1)), and the model says x(0) is True and x(1) is also True, then this clause becomes [1, 0].

Think step by step. Show the variable assignments and how you evaluate each clause.

Finally, in the **last line**, output a single line in the format:
Assignment: [[1, 0], [0, 1, 1], [1], ...]

For any variable that is not explicitly mentioned in the reasoning trace, assume its value is 0 when constructing the assignment list.

Do not include anything after this label.
\end{lstlisting}

\end{tcolorbox}
\caption{Trace Evaluation Prompt Template for SAT Prediction.}
\label{fig:trace-eval}
\end{figure*}

\begin{figure*}[t] 
\centering
\begin{tcolorbox}[title=Trace Evaluation Prompt Template for UNSAT Prediction,
  colback=gray!5, colframe=gray!50,
  boxrule=0.8pt, arc=4pt, outer arc=4pt, 
  width=\textwidth]  

\begin{lstlisting}[basicstyle=\ttfamily\small, breaklines=true, mathescape=true]
You are evaluating whether a model's reasoning trace correctly explains an UNSAT logical puzzle.

<scenario>
{scenario}

<conditions>
{conditions}

<question>
{question}

<variable explanation>
{variable_mapping}

<reasoning trace from model>
{model_trace}

<ground-truth unsat reason>
{unsat_reason}

We already know this puzzle is UNSAT (unsatisfiable).  
Your task is to judge whether the reasoning trace correctly identifies or meaningfully reflects the cause of unsatisfiability - that is, whether it aligns with the given ground-truth unsat reason, even if it doesn't name it explicitly.

Focus on logical precision:  
- Does the trace show or imply a variable assignment or chain of reasoning that leads to contradiction?  
- Does it avoid hallucinations or irrelevant claims?

Note: The trace may present a specific variable assignment or reasoning path that leads to a contradiction. Whether it aligns with the given ground-truth UNSAT reason means you must judge whether the contradiction is logically valid and reflective of the actual cause, even if it doesn't explicitly name the minimal core or unsat pattern.

You are **not** evaluating whether the conclusion "UNSAT" is correct - that is already known to be correct.  
You are only evaluating whether the explanation substantively captures why the instance is unsatisfiable.

Please think step by step. First, explain whether and how the reasoning trace aligns with the unsat reason.  
Then, in the last line, output one of the following labels:

[YES] - the reasoning trace is logically valid and correctly captures the UNSAT cause  
[NO] - the trace is flawed, incomplete, or does not match the correct unsat reason

Do not include anything after this label.
\end{lstlisting}

\end{tcolorbox}
\caption{Trace Evaluation Prompt Template for UNSAT Prediction.}
\label{fig:trace-eval2}
\end{figure*}

\section{Error Analysis Examples}
\label{app:error-examples}

We show one representative example for each of the four error types, paraphrased for clarity.

\paragraph{\textbf{Satisfiability Bias}} 
The model outputs an assignment such as $x(1)=1, x(2)=1, x(3)=0$ and prematurely declares the formula satisfiable. In reality, satisfiability requires that \emph{all} clauses be satisfied, yet several clauses remain violated. This suggests that the model often assumes satisfiability without exhaustively checking all constraints and fails to engage in search-based logical reasoning with backtracking.

\paragraph{\textbf{Context Inconsistency}} 
The model produces conflicting assignments for the same variable within one trace. For example, it first sets $x(0)=1$ but later assigns $x(0)=0$, trying to satisfy $x(0) \land \neg x(0)$. This is impossible: a variable can only take a single value. The correct resolution is either to retain one consistent assignment in a satisfiable case or to conclude UNSAT when no such assignment exists.

\paragraph{\textbf{Condition Omission}}  
The model may ignore or hallucinate conditions in its reasoning trace. For example, given \((x(0) \lor x(1)) \land \neg x(0)\), it incorrectly reduces the formula to \(x(0) \land \neg x(0)\), which is unsatisfiable. In reality, the original formula is satisfiable with \(x(0)=0, x(1)=1\). Such omissions cause the model to misclassify satisfiable instances as UNSAT.

\paragraph{\textbf{Spurious Priors}}  
The model introduces commonsense assumptions that are absent from the formula. For example, with $x(0) \lor x(1)$, the model assumes that $x(0)$ and $x(1)$ cannot both be true (as if they were “mutually exclusive”). It then treats the assignment $x(0)=1, x(1)=1$ as a contradiction and concludes UNSAT. In reality, the formula is satisfiable and permits both variables to be true. This is because models sometimes introduce commonsense assumptions that are absent from the given constraints.

\end{document}